\documentclass{article}

\usepackage{instructvvt_arxiv}

\usepackage[utf8]{inputenc}
\usepackage[T1]{fontenc}
\usepackage{float}
\usepackage{hyperref}
\hypersetup{hidelinks}
\usepackage{url}
\usepackage{booktabs}
\usepackage{amsfonts}
\usepackage{amsmath}
\usepackage{amssymb}
\usepackage{microtype}
\usepackage{xcolor}
\usepackage{graphicx}

\title{InstructVVT: Instruction-Driven Video Virtual Try-On without Auxiliary Spatial Priors}

\author{\normalsize
\textbf{Dingbao Shao}\textsuperscript{*},
\textbf{Song Wu}\textsuperscript{*},
\textbf{Xinyu Chen},
\textbf{Qian Wang},
\textbf{Jiahang Li},
\textbf{Kuai Jiang},
\textbf{Jiang Lin} \\
\textbf{Yuhang Liu},
\textbf{Ziyu Chen},
\textbf{Duo Li},
\textbf{Jiaxin Hu},
\textbf{Shengrong Gu},
\textbf{Ziheng Tang},
\textbf{Rongrong Liu} \\
\textbf{Yanlun Peng},
\textbf{Liang Li},
\textbf{Junlan Feng},
\textbf{Lujia Jin},
\textbf{Ting Zhang},
\textbf{Jian Yang},
\textbf{Zili Yi}\textsuperscript{\textdagger} \\[-0.25ex]
{\normalfont\small
* Equal contribution.\quad \textdagger\ Corresponding author.}
}

\begin{document}

\maketitle

\begin{abstract}
Video virtual try-on is a highly constrained editing task requiring the precise replacement of a target person's clothing while strictly preserving the original video's spatial structure and temporal dynamics. Existing methods heavily rely on auxiliary handcrafted spatial priors (e.g., masks, poses) for editing control. However, these priors are prone to failure in unconstrained real-world videos and often compress rich visual context into incomplete structural signals. Furthermore, standard reconstruction objectives fail to fully capture try-on-specific human preferences. To address these challenges, we propose InstructVVT, an instruction-driven and reference-guided video virtual try-on framework based on a Diffusion Transformer (DiT) that operates without inference-time spatial priors. Our core insight is to recover fine-grained control directly from the input triplet (source video, reference garment, and instruction) via a dual-level reference conditioning scheme. Specifically, an MLLM infers semantic edit tokens for target disambiguation and structural preservation, while a lightweight conditioning pathway explicitly injects fine-grained visual garment details. Finally, we design a try-on-specific reward and utilize the DiffusionNFT algorithm to align the model with human preferences. Extensive experiments on ViViD-S and TripVVT-Bench demonstrate that InstructVVT outperforms state-of-the-art open-source methods in garment fidelity, structural preservation, and temporal consistency, despite requiring fewer inference-time controls.
\end{abstract}

\begin{figure}[H]
\centering
\includegraphics[width=0.94\textwidth]{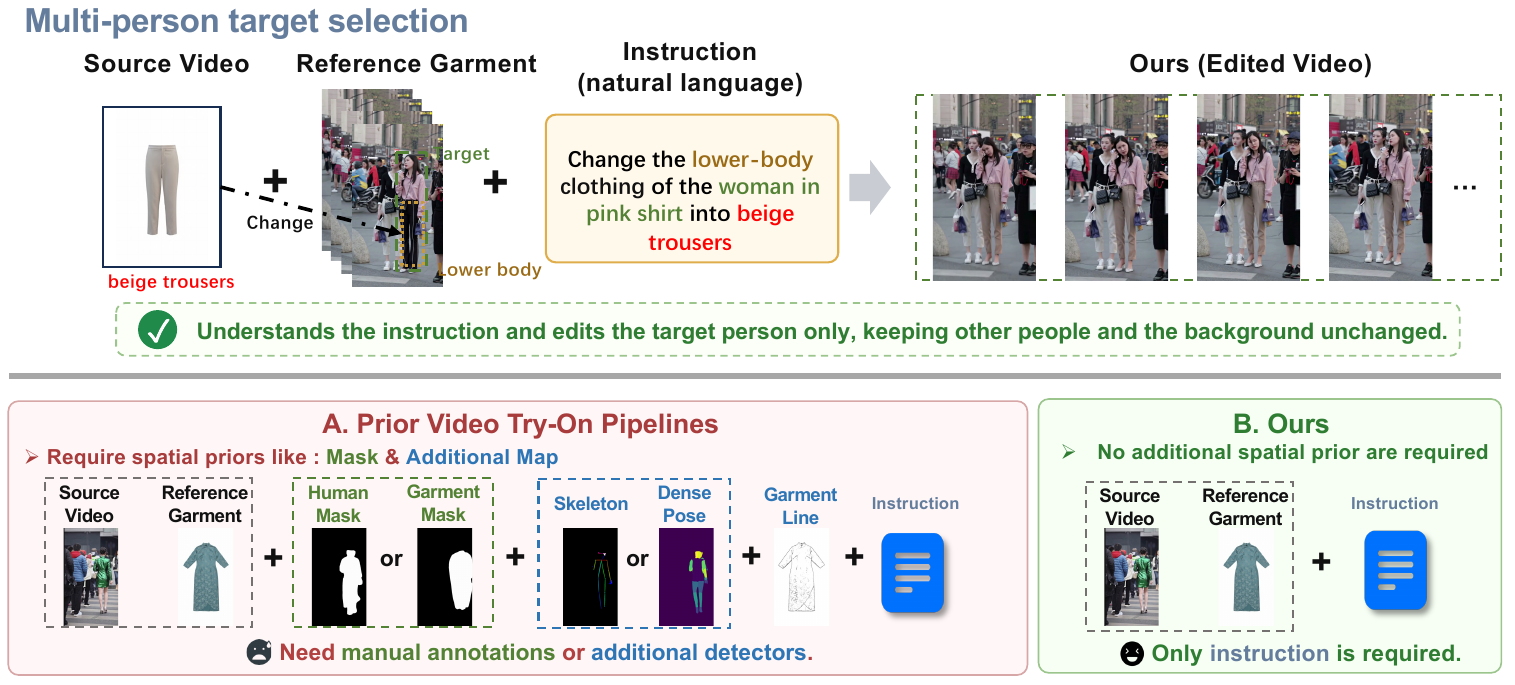}
\caption{Teaser examples of instruction-driven video virtual try-on. Given a source video, a reference garment, and a natural-language instruction, the model edits the instructed target person while preserving the source identity, motion, background, and temporal consistency.}
\label{fig:teaser}
\vspace{-0.3cm}  
\end{figure}

\section{Introduction}

Video virtual try-on constitutes a video editing task with reference constraints and structural preservation properties~\citep{vivid,dreamvvt,magictryon,tripvvt}. Taking a source video and a reference garment image as inputs, the task requires the model to replace the clothing of the target person, while maintaining the consistency of personal identity, body shape, pose dynamics, camera movement, scene layout, background content and temporal coherence. Different from general video generation tasks, visual realism is far from sufficient for video virtual try-on. Specifically, the edited video should faithfully restore the style and texture of the reference garment, accurately apply the clothing replacement to the designated human subject, and keep irrelevant scene content completely unchanged.

The core difficulty of video virtual try-on is to provide the model with precise editing control: it must know who to edit, where to edit, how to transfer the garment, and what to preserve. Existing methods typically obtain such control from auxiliary spatial priors~\citep{tryondiffusion,stableviton,idmvton,vivid,dreamvvt,magictryon,tripvvt}, including human masks, garment masks, parsing maps, DensePose representations, skeletons, or garment structure lines. These priors provide explicit localization and preservation cues, enabling high-quality try-on in constrained settings. However, three limitations stand out. (i) Their performance depends heavily on external estimators or manual preprocessing, which can be unreliable in unconstrained videos with unusual viewpoints, occlusions, loose garments, large camera motion, or multiple people. (ii) Since these controls are predefined and task-specific, they may compress the source video and reference garment into incomplete structural signals, losing useful visual context for target disambiguation, garment-to-body alignment, and non-edited region preservation. (iii) Supervised reconstruction or denoising objectives alone cannot fully capture the qualities that define a good try-on result, and may lead to weakened garment details, unintended changes to non-target regions, incorrect target editing, or temporal flicker. Recently, general instruction-guided video editing models have shown strong editing capabilities~\citep{dreamix,anyv2v,univideo}. Nevertheless, they are not specifically optimized for the fine-grained constraints of video virtual try-on, and often struggle with reference-garment fidelity, target correctness, background preservation, and temporal stability.

To address these challenges, we propose InstructVVT, a DiT-based video diffusion framework for instruction-driven and reference-guided video virtual try-on, as shown in Fig.~\ref{fig:method_overview}. InstructVVT takes a source video, a reference garment image, and a natural-language instruction as inputs, and generates an edited video in which the instructed person wears the reference garment. During inference, it does not require auxiliary handcrafted spatial priors. Instead, InstructVVT recovers editing control from the input triplet itself. Specifically, we use a multimodal large language model (MLLM) to jointly parse the source video, the reference garment, and the instruction, and extract garment-aware edit-intent tokens to guide the video diffusion model. Compared with handcrafted spatial priors, these edit-intent tokens provide a more flexible semantic interface: they help resolve the target subject, the editable clothing region, the preservation constraints, and the high-level alignment between the garment and the person. In parallel, source-video latents are fed into the generator to anchor motion, layout, camera behavior, and non-edited regions. To improve reference-garment fidelity, we further design a lightweight garment conditioning branch that injects fine-grained visual garment tokens into the Diffusion Transformer. This dual use of the reference garment is important: the MLLM provides semantic garment-to-person alignment, while the generator-side garment tokens preserve texture, pattern, shape, and visible design details. Together, these conditions allow the generator to treat video virtual try-on as a localized edit of the source video, rather than synthesis driven by externally specified spatial maps.

In addition, inspired by recent MLLM-based feedback methods~\citep{uniworldv2}, we design a training-free video try-on reward to address the limitation that supervised losses do not fully characterize try-on quality. The reward evaluates generated videos from try-on-specific aspects, including garment fidelity, instruction following, target correctness, identity and background preservation, motion preservation, and temporal consistency. We combine this reward with DiffusionNFT algorithm~\citep{diffusionnft} to align the generator with human try-on preferences.

Experiments on ViViD-S~\citep{catv2ton} and TripVVT-Bench~\citep{tripvvt} show that InstructVVT outperforms current open-source state-of-the-art video try-on models across evaluation metrics and human preference, while requiring no auxiliary spatial controls at inference time. Our contributions are threefold:
\begin{itemize}
  \item We propose InstructVVT, a DiT-based video diffusion framework that establishes an instruction-driven and reference-guided video virtual try-on paradigm without relying on inference-time auxiliary handcrafted spatial priors.
  \item We present a dual-level reference conditioning scheme for DiT-based video try-on: an MLLM jointly analyzes the source video, reference garment, and instruction for garment-aware edit planning, while a lightweight garment-token pathway provides fine-grained visual appearance details.
  \item We design a try-on-specific reward based on a frozen MLLM and use DiffusionNFT-based post-training to improve garment fidelity, target correctness, source preservation, motion naturalness, and temporal consistency.
\end{itemize}

\section{Related Work}

\subsection{Virtual try-on with spatial priors}

Virtual try-on aims to transfer a reference garment to a target person while preserving identity, pose, body shape, and background content. Image virtual try-on has been studied extensively with warping-based, parsing-based, and diffusion-based frameworks~\citep{tryondiffusion,stableviton,idmvton,catvton}, while video virtual try-on further requires temporal consistency under body motion, camera movement, and occlusion~\citep{vivid,catv2ton,dreamvvt,magictryon,tripvvt}. To achieve controllable editing, many existing methods rely on explicit spatial priors, such as human masks, garment masks, parsing maps, poses, DensePose-like representations, skeletons, or garment structure lines. These signals provide useful localization and preservation cues, but also introduce dependence on external estimators or manual preprocessing. In challenging in-the-wild videos, such priors may be inaccurate, temporally unstable, or unavailable.

Recent mask-free or reduced-condition try-on methods improve usability by removing some spatial inputs, especially garment or edit-region masks~\citep{pemfvto,boowvton,mfpvton,mfviton}. However, they often still depend on other structural conditions or assume a relatively fixed try-on interface. This makes it difficult to handle natural-language instructions that specify the target person, edit scope, and preservation requirements jointly. In contrast, our method removes inference-time auxiliary structural priors and uses only a source video, a reference garment, and an instruction.

\subsection{Instruction-guided and multimodal visual editing}

Instruction-guided image and video editing methods provide a more flexible interface than task-specific controls~\citep{instructpix2pix,magicbrush,dreamix,anyv2v}. By conditioning generation on natural-language commands, these models can perform diverse edits without manually specifying low-level spatial annotations. More recently, MLLM-guided editing systems use multimodal representations to interpret visual context together with the user instruction~\citep{mgie,metaquery,instructx,univideo}, which is especially useful when the target object is described by attributes, relations, or scene context rather than by an explicit mask.

However, video virtual try-on has stricter constraints than generic visual editing. A successful try-on result must preserve the source identity, body motion, scene layout, and non-target regions, while faithfully transferring the reference garment and maintaining temporal stability. General editing models may follow the instructions but drift from the garment reference, edit the wrong person, alter the background, or introduce inconsistent clothing details across frames. Our work adapts MLLM-based conditioning to video try-on by using the MLLM to jointly parse the source video, reference garment, and instruction into edit-intent tokens, while also injecting the garment appearance through a generator-side visual branch.

\subsection{Reward alignment for diffusion models}

Most diffusion-based editing and try-on models are trained with reconstruction, denoising, or flow-matching objectives. These objectives are effective for learning paired data distributions, but they do not directly optimize human preferences. In virtual try-on, important criteria such as garment fidelity, target correctness, source preservation, visual naturalness, and temporal consistency are difficult to capture with a single supervised loss.

Reward-based post-training provides a way to align diffusion models with such high-level preferences~\citep{dpok,ddpo,diffusiondpo,diffusionnft}. Instead of relying only on pixel-level or feature-level reconstruction, a reward model can evaluate generated samples according to task-specific criteria and guide the generator toward preferred outputs. Recent instruction-based editing work also explores training-free MLLM implicit feedback as a reward signal~\citep{uniworldv2}. In this work, we construct a training-free MLLM score-token reward for video try-on and use DiffusionNFT-based post-training to improve try-on-specific qualities after supervised fine-tuning.

\section{Method}
\label{sec:method}

\subsection{Overview}
\label{sec:overview}

\begin{figure*}[t]
\centering
\includegraphics[width=\textwidth]{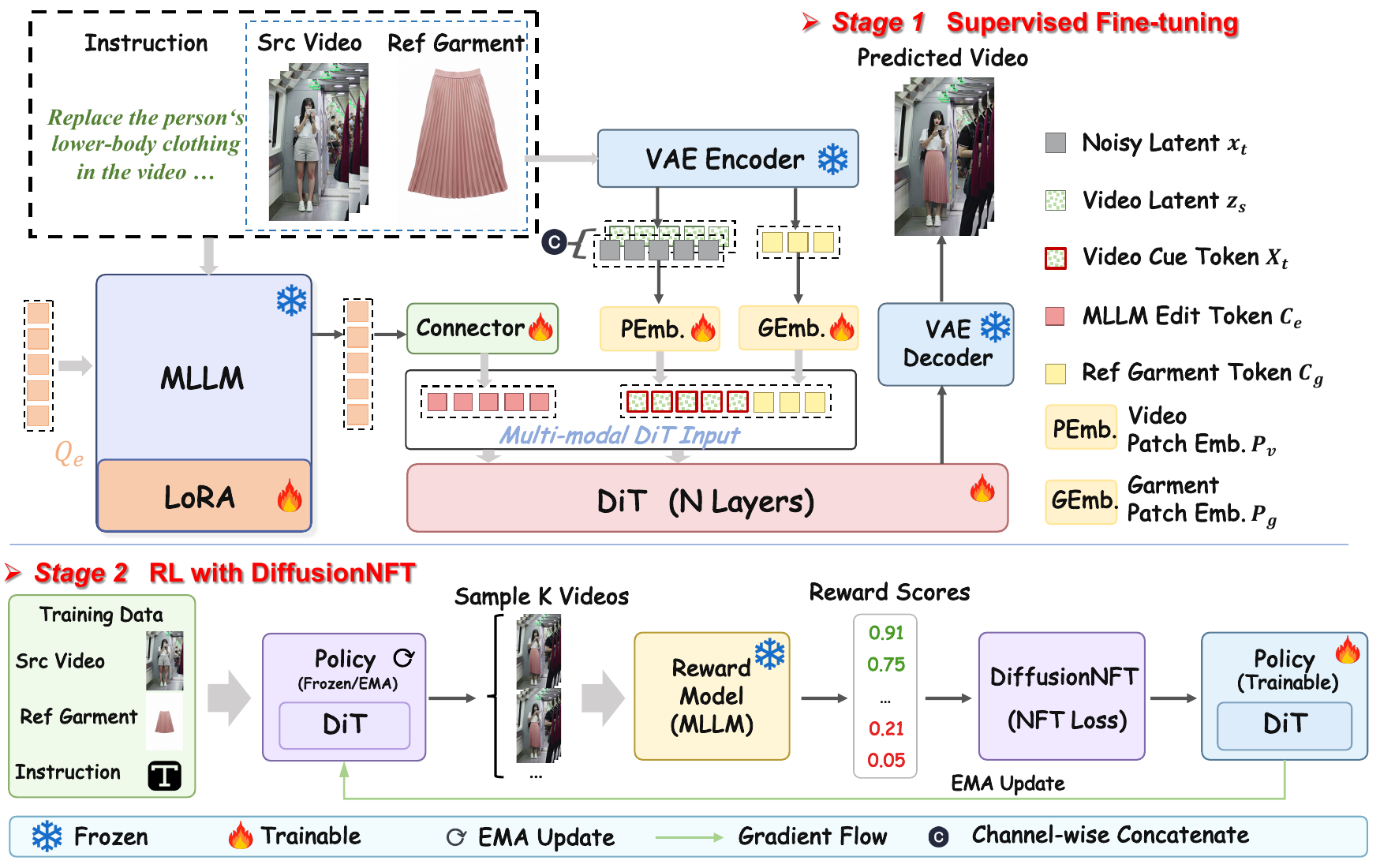}
\caption{
Overview of our InstructVVT framework.
In the supervised training, the source video, reference garment, and instruction are encoded by a frozen MLLM with trainable query tokens $Q_e$, LoRA adapters~\citep{lora}, and an MLP connector to produce MLLM edit tokens $C_e$. We adopt two embedding layers with identical architecture but independent parameters to encode video cue tokens $X_t$ and reference garment tokens $C_g$, respectively. The MLLM edit tokens $C_e$, video cue tokens $X_t$ and reference garment tokens $C_g$ are injected into each DiT block through cross-attention. The post-training stage applies DiffusionNFT~\citep{diffusionnft}: a frozen EMA policy samples $K$ videos, a tailored reward model based on MLLM assigns score-token rewards, and the NFT loss updates the trainable DiT.
}
\label{fig:method_overview}
\end{figure*}

As shown in Fig.~\ref{fig:method_overview}, our framework is built on an open source latent video Diffusion Transformer (DiT)~\citep{wan,dit} and an MLLM. Given a source video $V_{\mathrm{s}}$, a reference garment image $I_{\mathrm{g}}$, and a natural-language instruction $\mathcal{I}$, our model generates an edited video $\hat{V}$ in which the instructed person wears the reference garment while the remaining content is preserved. At inference time, our model uses only these three inputs and does not require auxiliary structural annotations such as masks, poses, parsing maps, DensePose-like conditions~\citep{densepose}, skeletons, or garment contours.

Our model comprises three complementary conditioning pathways. First, latents encoded from the source video by a pretrained VAE preserve motion, layout, and non-edited regions. Second, a frozen MLLM with trainable LoRA parameters~\citep{lora} takes sampled frames from the source video, reference garment and instruction as inputs. It generates garment-aware edit tokens to locate the target subject, define the edit scope, and establish high-level alignment between the garment and the person. Third, reference-garment tokens provide explicit visual appearance via a generator-side pathway. The latter two garment-aware pathways are complementary: the MLLM branch enhances semantic alignment with the reference garment, while the generator-side garment tokens retain fine-grained texture, pattern, shape and subtle design details.

We adopt two major training paradigms: supervised fine-tuning and reinforcement learning training. The supervised fine-tuning paradigm learns the correspondence among source video, reference garment, instruction, and target try-on video. The reinforcement learning stage employs a frozen MLLM~\citep{qwen3vl} as the reward model and leverages the DiffusionNFT~\citep{diffusionnft} algorithm to align the generator with specific preferences for virtual try-on. Experimental details, stage-wise training settings, and compute resources are provided in Appendix~\ref{app:training_details}.

\subsection{Instruction-Driven Video Virtual Try-On (InstructVVT)}
\label{sec:generator}

\paragraph{MLLM Edit Tokens.}
A text-only instruction is often insufficient for video try-on. As shown in Fig.~\ref{fig:teaser}, instructions like ``change the lower-body clothing of the woman in the pink shirt to beige trousers.'' require the model to understand the source video. It needs to detect all people, infer their spatial relations, locate clothing regions, and identify content that should stay unchanged. Moreover, a single text instruction cannot fully describe the fine-grained garment appearance, nor clarify how the reference garment aligns with the target body. We thus leverage an MLLM to jointly process sampled source frames, the reference garment image, and the instruction $\mathcal{I}$. The MLLM infers the editing context, including the target person, editing scope, garment-body alignment, and content preservation constraints.

Following recent methods~\citep{metaquery,instructx}, we introduce learnable query tokens $Q_{\mathrm{e}}$ to extract compact editing features from the MLLM. The query tokens are appended to the multimodal prompt and are processed together with the source frames, reference garment, and instruction. We then select the hidden states at the query-token positions and project them to the DiT hidden dimension:
\begin{equation}
C_{\mathrm{e}}
=
P_{\mathrm{e}}
\left(
\operatorname{Select}_{Q}
\left[
\operatorname{MLLM}(V_{\mathrm{s}}, I_{\mathrm{g}}, \mathcal{I}, Q_{\mathrm{e}})
\right]
\right),
\label{eq:edit_tokens}
\end{equation}
where $P_{\mathrm{e}}$ is an MLP connector and $C_{\mathrm{e}}$ denotes the MLLM edit tokens.

Since the MLLM observes the reference garment, the edit tokens encode not only source-aware target localization and edit scope, but also high-level garment semantics useful for aligning the reference clothing to the target person. These tokens provide semantic and relational guidance rather than explicit pixel-level masks, poses, parsing maps, or edit regions.

\paragraph{Video Cue Tokens and Garment Tokens.}
Let $x_t$ be the noisy video latent at diffusion timestep $t$, and let
$z_{\mathrm{s}}=\mathcal{E}_{\mathrm{vae}}(V_{\mathrm{s}})$ be the source-video latent. The source latent $z_t$ is concatenated with the noisy latent $x_t$ along the channel dimension and fed into the video patch embedding layer to yield video cue tokens $X_t$:
\begin{equation}
X_t
=
P_{\mathrm{v}}
\left(
\operatorname{Concat}_{\mathrm{ch}}(x_t,z_{\mathrm{s}})
\right),
\label{eq:source_tokens}
\end{equation}
where $P_{\mathrm{v}}$ is the video patch embedding layer. Since the pretrained patch embedding expects only target-latent channels, we expand its input channel dimension by a factor of two. The original channel weights are initialized from the pretrained model, and the newly added source-channel weights are initialized to zero for stable fine-tuning.

The reference garment is encoded by the VAE and mapped into garment tokens $C_{\mathrm{g}}$ via an independent garment patch embedding layer:
\begin{equation}
C_{\mathrm{g}}
=
\operatorname{Flatten}
\left(
P_{\mathrm{g}}
\left(
\mathcal{E}_{\mathrm{vae}}(I_{\mathrm{g}})
\right)
\right),
\label{eq:garment_tokens}
\end{equation}
where $P_{\mathrm{g}}$ denotes the garment patch embedding layer and $C_{\mathrm{g}}$ denotes the reference-garment tokens. This generator-side garment pathway provides fine-grained visual evidence for synthesis and does not use garment masks, contours, parsing maps, or structure-line extractors.

\paragraph{Condition Injection within DiT.}
The video tokens $X_t$ and garment tokens $C_{\mathrm{g}}$ are concatenated along the token dimension before entering the DiT:
\begin{equation}
H_0
=
\operatorname{Concat}_{L}
\left(
X_t,
C_{\mathrm{g}}
\right).
\label{eq:token_concat}
\end{equation}
We replace the original T5 text embeddings of the base DiT~\citep{wan} with our MLLM edit tokens $C_{\mathrm{e}}$. These tokens are injected into each DiT block via cross-attention:
\begin{equation}
H_{\ell+1}
=
\operatorname{DiTBlock}_{\ell}
\left(
H_{\ell}, t, C_{\mathrm{e}}
\right),
\qquad
\ell=0,\ldots,L_{\mathrm{dit}}-1.
\label{eq:dit_block}
\end{equation}
After the final DiT block, only the video-token positions are unpatchified to predict the target latent; the garment tokens serve as conditioning tokens and are discarded before VAE decoding.

This design lets the model recover try-on controllability from the input triplet itself. The MLLM edit tokens provide instruction- and reference-aware editing context, the source latent provides dense structural anchoring, and the garment tokens provide explicit appearance cues, all without inference-time auxiliary spatial priors.

\subsection{Two major training paradigms}

\paragraph{Supervised Fine-Tuning.}
Given a paired example $(V_{\mathrm{s}}, I_{\mathrm{g}}, \mathcal{I}, V_{\mathrm{t}})$, we encode the target video as $x_0=\mathcal{E}_{\mathrm{vae}}(V_{\mathrm{t}})$ and optimize the standard flow-matching objective~\citep{flowmatching}:
\begin{equation}
\mathcal{L}_{\mathrm{SFT}}
=
\mathbb{E}_{t}
\left[
\left\|
f_{\theta}
\left(
x_t,z_{\mathrm{s}},C_{\mathrm{g}},C_{\mathrm{e}},t
\right)
-
v
\right\|_2^2
\right],
\label{eq:sft_loss}
\end{equation}
where $x_t$ and $v$ are produced by the flow-matching noise schedule. This stage teaches the model the basic correspondence among the source video, reference garment, instruction, and target try-on video. In this stage, the MLLM backbone is kept frozen, while the query-token embeddings, LoRA adapters, and connector are trainable. 


\paragraph{Reinforcement Learning with Try-On-Specific Reward.} 
\label{sec:reward_alignment}

Supervised fine-tuning does not directly optimize the preference criteria that matter for try-on, such as reference-garment fidelity, correct target editing, preservation of non-edited regions, natural motion, and temporal stability. We therefore use a frozen pretrained MLLM as a training-free reward model. This reward MLLM is separate from the LoRA-adapted MLLM used for generator conditioning.

Given $(V_{\mathrm{s}}, I_{\mathrm{g}}, \mathcal{I}, \hat{V})$, the reward MLLM is prompted with a fixed try-on rubric and asked to directly output one score from $0$ to $5$. Instead of parsing generated text, we compute a score-token reward from the first-token logits. Let $\ell_j$ be the logit of score token $j\in\{0,1,2,3,4,5\}$. We define
\begin{equation}
p_j
=
\frac{\exp(\ell_j)}
{\sum_{m=0}^{5}\exp(\ell_m)}\qquad \text{and}
\qquad
\rho(\hat{V},c)
=
\frac{1}{5}
\sum_{j=0}^{5}
j\,p_j,
\label{eq:score_logit_reward}
\end{equation}
where $c=(V_{\mathrm{s}},I_{\mathrm{g}},\mathcal{I})$ and $\rho\in[0,1]$. This non-CoT score-logit reward avoids brittle text parsing and provides a smooth scalar signal for post-training.

For each condition $c$, a frozen EMA policy samples candidate videos, which are scored by the reward model. We then apply DiffusionNFT~\citep{diffusionnft} to update the trainable generator. Let $r=\rho(\hat{V},c)$ be the reward of a sampled video. For its latent $x_0$, we draw a timestep $t$ and compute the noised latent $x_t$ and velocity target $v$ using the same flow-matching schedule. Let $v_{\theta_{\mathrm{old}}}$ and $v_{\theta}$ denote the EMA and trainable velocity predictors. DiffusionNFT constructs
\begin{align}
v_{\theta}^{+}(x_t,c,t)
&=
(1-\beta)
v_{\theta_{\mathrm{old}}}(x_t,c,t)
+
\beta
v_{\theta}(x_t,c,t)
\end{align}
and
\begin{align}
v_{\theta}^{-}(x_t,c,t)
&=
(1+\beta)
v_{\theta_{\mathrm{old}}}(x_t,c,t)
-
\beta
v_{\theta}(x_t,c,t).
\end{align}
The post-training loss is
\begin{equation}
\mathcal{L}_{\mathrm{NFT}}
=
\mathbb{E}
\left[
r
\left\|
v_{\theta}^{+}(x_t,c,t)-v
\right\|_2^2
+
(1-r)
\left\|
v_{\theta}^{-}(x_t,c,t)-v
\right\|_2^2
\right].
\label{eq:nft_loss}
\end{equation}
High-reward samples guide the generator toward preferred try-on behavior, while low-reward samples define directions to avoid. The reward model is used only during post-training and is not required at inference time.
\section{Experiments}
\label{sec:experiments}

\subsection{Experimental Setup}
\label{sec:exp_setup}

\paragraph{Training data and protocol.}
We train InstructVVT on source--reference--target try-on examples from both in-the-wild and indoor domains. The corpus consists of TripVVT-10K~\citep{tripvvt}, 10K supplementary ViViD~\citep{vivid} video triplets, and around 100K image triplets constructed from public try-on datasets including VITON-HD~\citep{vitonhd} and DressCode~\citep{dresscode}; all evaluation videos and garments are excluded from training. Training follows supervised try-on learning followed by DiffusionNFT post-training with the proposed MLLM reward. Appendix~\ref{app:training_data_construction} details data construction, Table~\ref{tab:experimental_setting_details} reports stage-wise training and compute settings, and Appendix~\ref{app:reward_details} gives the reward-model details and RL reward curve in Fig.~\ref{fig:rl_reward_curve}.

\paragraph{Evaluation benchmarks and metrics.}
We evaluate on ViViD-S test, following CatV2TON~\citep{catv2ton}, and TripVVT-Bench~\citep{tripvvt}, which contains more challenging in-the-wild videos with complex motion, diverse scenes, and multi-person cases. Following the TripVVT-Bench protocol, we report metrics covering video quality (VFID$_I$, VFID$_R$), reference-garment fidelity (CLIP-I), background consistency (BG-L1, BG-DINO), temporal consistency (CLIP-F), and reconstruction fidelity when target videos are available (SSIM, LPIPS)~\citep{fid,fvd,i3d,resnext3d,clip,dino,ssim,lpips}. We additionally report MLLM-Reward because Sec.~\ref{sec:human_reward_eval} shows its positive correlation with human preference, and its open-source backbone makes the evaluation easier to reproduce. Table~\ref{tab:evaluation_settings} provides benchmark sizes and inference settings, while Appendix~\ref{app:stage_training_settings} describes the reproducibility artifacts.

\paragraph{Baselines.}
We compare with open-source video virtual try-on models, including ViViD~\citep{vivid}, CatV2TON~\citep{catv2ton}, MagicTryOn~\citep{magictryon}, and TripVVT~\citep{tripvvt}, as well as the open-source general video editing model UniVideo~\citep{univideo}. On TripVVT-Bench, we additionally include commercial video editing models, including Kling 1.6~\citep{kling_1.6} and Kling O3~\citep{kling_o3}, to evaluate against strong general-purpose editors. For methods that require auxiliary spatial priors, we provide the required conditions through their original preprocessing pipelines; Kling 1.6 is evaluated with its available point/region-based editing interface, while Kling O3 uses its video-and-instruction editing interface without region-level control. Additional baseline protocol details are given in Appendix~\ref{app:stage_training_settings}.
\subsection{Quantitative Comparisons}
\label{sec:quantitative}

\paragraph{Results on ViViD-S test.}
Table~\ref{tab:vivid_s} reports quantitative results on ViViD-S test. InstructVVT achieves the best VFID$_I$, VFID$_R$, background metrics, and MLLM-Reward, while maintaining near-best temporal consistency. The discrepancy between generic metrics and MLLM-Reward further shows that video smoothness or generic editing ability alone does not ensure reference-faithful, source-preserving video try-on.
\begin{table*}[t]
\centering
\caption{Quantitative comparison with other methods on ViViD-S~\citep{catv2ton} test dataset. We additionally report the proposed MLLM-Reward score. Best results are shown in \textbf{bold}, and second-best results are \underline{underlined}.}
\label{tab:vivid_s}
\resizebox{\textwidth}{!}{
\begin{tabular}{@{}l c c c c c c@{}}
\toprule
Method & VFID$_I$ $\downarrow$ & VFID$_R$ $\downarrow$ & CLIP-F $\uparrow$ & BG-L1 $\downarrow$ & BG-DINO $\downarrow$ & MLLM-Reward $\uparrow$ \\
\midrule
ViViD~\citep{vivid} & 21.8032 & 0.8212 & 0.9574 & 0.0571 & \underline{0.0055} & 0.6341 \\
CatV2TON~\citep{catv2ton} & 19.5131 & 0.5283 & 0.9304 & 0.0507 & 0.0071 & 0.5773 \\
MagicTryOn~\citep{magictryon} & 17.5710 & 0.5073 & 0.9647 & \underline{0.0493} & 0.0056 & 0.5199 \\
TripVVT~\citep{tripvvt} & \underline{16.4398} & \underline{0.3315} & 0.9670 & 0.0624 & 0.0059 & 0.6034 \\
UniVideo~\citep{univideo} & 23.6970 & 1.2139 & \textbf{0.9965} & 0.1447 & 0.0124 & \underline{0.7152} \\
\midrule
Ours & \textbf{16.0271} & \textbf{0.2809} & \underline{0.9957} & \textbf{0.0267} & \textbf{0.0048} & \textbf{0.7236} \\
\bottomrule
\end{tabular}
}
\end{table*}

\paragraph{Results on TripVVT-Bench.}
Table~\ref{tab:tripvvt_bench} shows results on the more challenging TripVVT-Bench, which contains complex motion, camera changes, and possible target ambiguity. InstructVVT improves the main try-on, reconstruction, garment-fidelity, and human-preference metrics. General editors show complementary limitations: UniVideo is smooth but weak in source preservation, Kling 1.6 benefits from region-level hints but still trails in garment fidelity and reconstruction quality, and Kling O3 tends to regenerate rather than locally edit the source video.

\begin{table*}[t]
  \caption{Quantitative comparison with other methods on TripVVT-Bench. User preference reports first-place vote ratios from the four-method shortlist study; methods outside the shortlist are marked by ``--''. Best results are shown in \textbf{bold}, and second-best results are \underline{underlined}; ties share the same rank, with the next distinct value marked as second-best when applicable.}
  \label{tab:tripvvt_bench}
  \centering
  \resizebox{\textwidth}{!}{%
  \begin{tabular}{@{}l c c c c c c c c c c@{}}
    \toprule
    Method
    & VFID$_I \downarrow$
    & VFID$_R \downarrow$
    & SSIM $\uparrow$
    & LPIPS $\downarrow$
    & CLIP-I $\uparrow$
    & CLIP-F $\uparrow$
    & BG-L1 $\downarrow$
    & BG-DINO $\downarrow$
    & MLLM-Reward $\uparrow$
    & User Pref. $\uparrow$ \\
    \midrule
    ViViD~\citep{vivid}
    & 26.7620 & 0.7083 & 0.8323 & 0.1343 & 0.8716 & 0.9849 & 0.0691 & \underline{0.0053} & 0.5569 & -- \\

    CatV2TON~\citep{catv2ton}
    & 34.2275 & 3.3266 & 0.7845 & 0.2130 & 0.8077 & 0.9819 & 0.1032 & 0.0109 & 0.2811 & -- \\

    MagicTryOn~\citep{magictryon}
    & 22.9238 & 0.5502 & \underline{0.8641} & 0.1150 & 0.9102 & \underline{0.9881} & \underline{0.0515} & \textbf{0.0036} & 0.5979 & 5.4\% \\

    TripVVT~\citep{tripvvt}
    & \underline{20.7245} & \underline{0.3163} & 0.8538 & \underline{0.1053} & \underline{0.9373} & 0.9876 & 0.0852 & 0.0059 & 0.6330 & -- \\

    UniVideo~\citep{univideo}
    & 28.4413 & 3.0811 & 0.5461 & 0.1871 & 0.9172 & \textbf{0.9897} & 0.2659 & 0.0091 & 0.6915 & 1.6\% \\

    Kling 1.6~\citep{kling_1.6}
    & 22.4242 & 0.6289 & 0.6662 & 0.1353 & 0.9311 & 0.9504 & 0.1662 & 0.0064 & \textbf{0.7227} & \underline{13.6\%} \\

    Kling O3~\citep{kling_o3}
    & 32.0901 & 2.4129 & 0.7284 & 0.1933 & 0.9089 & 0.9863 & 0.2040 & 0.0094 & 0.5577 & -- \\

    \midrule
    Ours
    & \textbf{12.8589} & \textbf{0.1404} & \textbf{0.8742} & \textbf{0.0615} & \textbf{0.9528} & \underline{0.9881} & \textbf{0.0468} & \underline{0.0047} & \underline{0.7193} & \textbf{79.4\%} \\
    \bottomrule
  \end{tabular}%
  }
\end{table*}

\subsection{Human and Reward Evaluation}
\label{sec:human_reward_eval}

\paragraph{Human preference study.}
We use a four-method shortlist consisting of Ours, MagicTryOn, UniVideo, and Kling 1.6. We sample 50 TripVVT-Bench examples and assign them to 20 participants, yielding 1,000 votes in total. For each example, participants view the anonymized outputs in randomized order and rank the four results according to overall video quality, reference-garment fidelity, source preservation, background consistency, and temporal coherence. We count the first-place votes for each method and report the resulting first-place vote ratio in Table~\ref{tab:tripvvt_bench}.

\paragraph{Reward-human agreement.}
We further evaluate whether the proposed reward reflects human preference. We sample 100 TripVVT-Bench test examples, generate four videos for each example, and ask human annotators to rank the results according to overall try-on quality. The reward model scores the same videos with the fixed try-on rubric, and we compare the reward-induced ranking with human ranking using pairwise agreement and rank-correlation metrics. As shown in Table~\ref{tab:human_reward}, the proposed MLLM reward has positive agreement with human preference, suggesting that it provides a meaningful training signal for try-on-specific post-training. Appendix~\ref{app:reward_details} defines the agreement metrics and gives additional reward-model details.

\begin{table}[t]
\centering
\caption{Reward-human agreement on TripVVT-Bench. Agreement is computed on 100 test samples with four generated videos per sample.}
\label{tab:human_reward}
\small
\setlength{\tabcolsep}{4pt}
\begin{tabular*}{\linewidth}{@{\extracolsep{\fill}}lccc@{}}
\toprule
Metric & Pairwise agreement $\uparrow$ & Spearman $\rho$ $\uparrow$ & Kendall $\tau$ $\uparrow$ \\
\midrule
Ours MLLM reward & 65.5\% & 0.368 & 0.310 \\
\bottomrule
\end{tabular*}
\end{table}

\subsection{Ablation Studies}
\label{sec:ablation}

We conduct ablations on TripVVT-Bench to verify the roles of MLLM conditioning, garment-aware reasoning, source anchoring, generator-side garment tokens, and DiffusionNFT post-training. The variants replace the MLLM with the original T5 encoder, remove the reference garment from the MLLM input, remove the source-video condition, remove generator-side reference-garment tokens, or use only the supervised model without post-training.

Table~\ref{tab:ablation} shows that all components contribute to the final model. Replacing the MLLM with T5 causes the largest drop, while the two garment pathways are complementary: the MLLM garment input supports high-level alignment, and garment tokens preserve fine-grained appearance. Source-video conditioning improves preservation and temporal stability, and DiffusionNFT post-training improves both MLLM-Reward and independent quality metrics.

\begin{table*}[t]
\centering
\caption{Ablation studies on TripVVT-Bench. We use the same metric setting as the TripVVT-Bench quantitative comparison. Best results are shown in \textbf{bold}, and second-best results are \underline{underlined}; ties share the same rank, with the next distinct value marked as second-best when applicable.}
\label{tab:ablation}
\resizebox{\textwidth}{!}{
\begin{tabular}{@{}l c c c c c c c c c@{}}
\toprule
Variant
& VFID$_I \downarrow$
& VFID$_R \downarrow$
& SSIM $\uparrow$
& LPIPS $\downarrow$
& CLIP-I $\uparrow$
& CLIP-F $\uparrow$
& BG-L1 $\downarrow$
& BG-DINO $\downarrow$
& MLLM-Reward $\uparrow$ \\
\midrule
w/o MLLM, using T5 & 18.3755 & 0.3496 & 0.8467 & 0.0808 & 0.9320 & \textbf{0.9882} & 0.0584 & 0.0053 & 0.6581 \\
w/o garment input to MLLM & \underline{13.0479} & \textbf{0.1161} & \underline{0.8717} & \underline{0.0633} & \underline{0.9514} & \underline{0.9881} & \underline{0.0487} & \underline{0.0048} & 0.7064 \\
w/o source-video condition & 14.5686 & 0.1938 & 0.8659 & 0.0673 & 0.9476 & 0.9878 & 0.0538 & 0.0049 & \underline{0.7068} \\
w/o reference-garment tokens & 14.8048 & 0.1996 & 0.8681 & 0.0675 & 0.9438 & \underline{0.9881} & 0.0520 & \underline{0.0048} & 0.6321 \\
w/o DiffusionNFT post-training & 13.8390 & \underline{0.1381} & 0.8688 & 0.0656 & 0.9482 & \textbf{0.9882} & 0.0490 & \underline{0.0048} & 0.6859 \\
\midrule
Full model & \textbf{12.8589} & 0.1404 & \textbf{0.8742} & \textbf{0.0615} & \textbf{0.9528} & \underline{0.9881} & \textbf{0.0468} & \textbf{0.0047} & \textbf{0.7193}  \\
\bottomrule
\end{tabular}
}
\end{table*}

\subsection{Qualitative Results and Generalization}
\label{sec:qualitative}

\begin{figure*}[t]
\centering
\includegraphics[width=\textwidth]{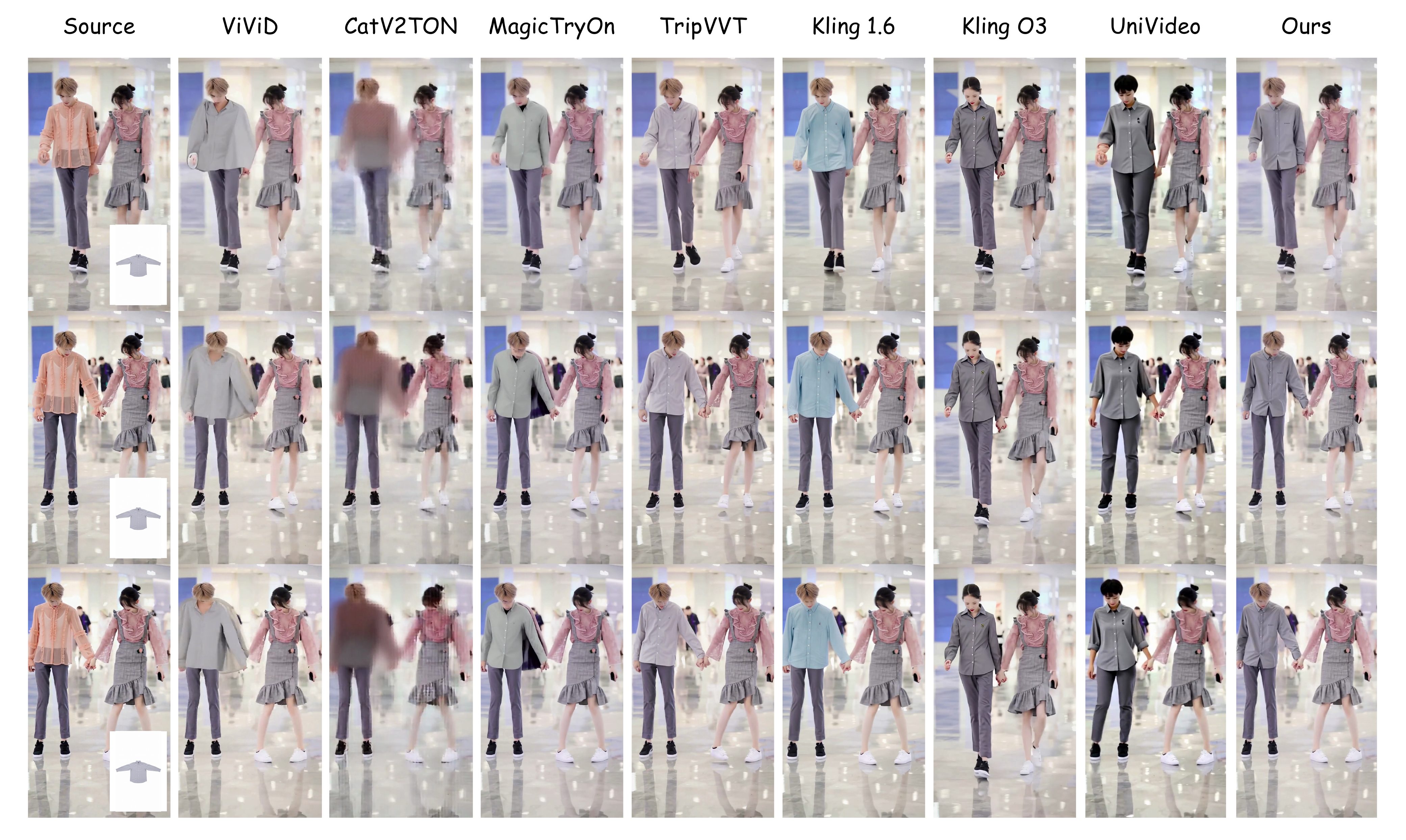}
\vspace{-0.5cm}  
\caption{Qualitative results on TripVVT-Bench. Our method preserves the source video structure and transfers the reference garment to the instructed subject with better temporal consistency.}
\label{fig:qualitative_main}
\vspace{-0.5cm}
\end{figure*}

\paragraph{Results on TripVVT-Bench.}
Fig.~\ref{fig:qualitative_main} shows that, on TripVVT-Bench, our method better preserves source structure and non-target regions while transferring the reference garment, especially under camera motion, occlusion, and multi-person ambiguity. It also produces more reference-faithful and temporally stable garment details than general video editors.

\paragraph{Results on ViViD-S.}
Additional ViViD-S examples in Appendix~\ref{app:additional_qualitative_results}, Figs.~\ref{fig:additional_qualitative_vivid_s} and~\ref{fig:additional_qualitative_vivid_s_more}, further show stable source preservation and reference-garment transfer on controlled indoor videos, complementing Table~\ref{tab:vivid_s}.

\section{Conclusion}

We presented InstructVVT, an instruction-driven framework for reference-based video virtual try-on that uses only a source video, a reference garment, and a natural-language instruction at inference time. Instead of relying on masks, poses, parsing maps, DensePose-like representations, garment lines, or other auxiliary spatial priors, the framework decomposes try-on control into three complementary signals: MLLM edit tokens for target-aware and garment-aware intent understanding, source-video latents for structure and motion anchoring, and generator-side garment tokens for fine-grained reference appearance preservation. We further introduced a try-on-specific MLLM reward and DiffusionNFT-based post-training to align the generator with garment fidelity, target correctness, source preservation, motion naturalness, and temporal consistency. Experiments on multiple video try-on benchmarks, together with systematic ablations, show that this design provides a practical and effective path toward controllable in-the-wild video try-on without inference-time auxiliary structural inputs.

\bibliographystyle{plainnat}
\bibliography{refs}
\clearpage
\appendix

\section{Limitations}
\label{app:limitations}

\begin{figure}[!t]
\centering
\includegraphics[width=0.92\linewidth]{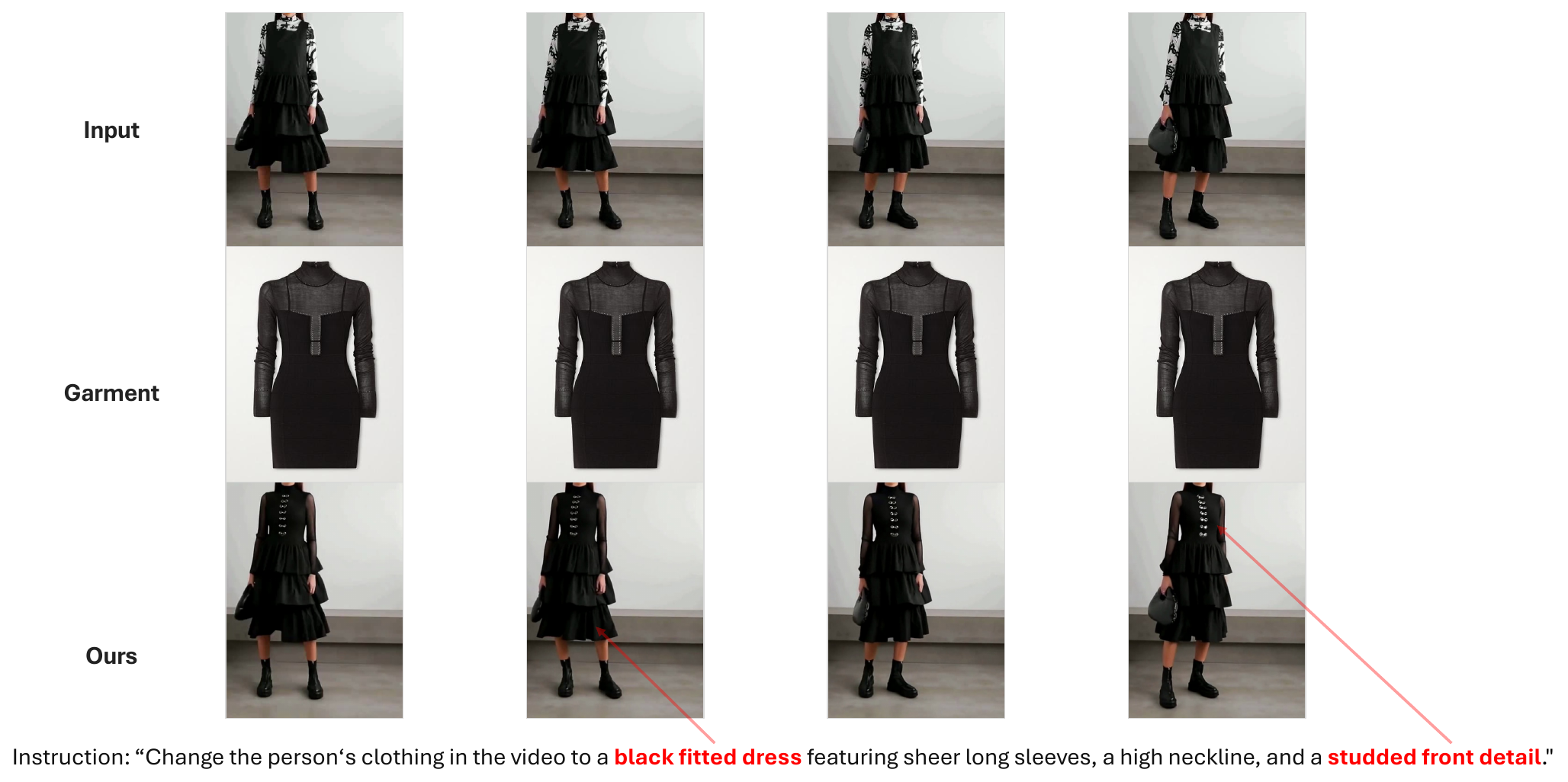}
\caption{A limitation case under conflict between text semantics and the reference garment image. When the instruction describes garment attributes that are inconsistent with the visual reference, the model may not fully preserve the reference garment appearance.}
\label{fig:limitation}
\end{figure}

The current design still depends on the visual reasoning ability of the MLLM and the quality of the reward model. Extremely long videos, severe occlusion, unusual garment geometry, and instructions requiring large physical changes may remain challenging. We also observe a failure mode when the text instruction semantically conflicts with the reference garment image. For example, if the instruction describes a garment attribute that is inconsistent with the reference image, the model may compromise between the text semantics and the visual reference, and the generated clothing may not fully follow the reference garment, as illustrated in Fig.~\ref{fig:limitation}.

Future work can improve conflict resolution between language and visual garment evidence, strengthen long-horizon consistency, expand reward supervision, and study more interactive forms of user control.

\section{Architecture details}
\label{app:architecture_details}

\paragraph{Architecture-specific settings.}
We uniformly sample four source frames for MLLM conditioning. The selected query-token hidden states are projected to the DiT cross-attention dimension by a two-layer MLP connector with GELU activation. The garment patch embedding is randomly initialized and trained from scratch. The resulting garment tokens participate in DiT self-attention as conditioning tokens, but are not decoded as output tokens. During DiffusionNFT post-training, the conditioning MLLM is frozen and only the video diffusion generator is updated; the reward MLLM is a separate frozen model.

\paragraph{Inference.}
At inference time, the model samples four source frames and feeds them, together with the reference garment and instruction, to the conditioning MLLM to obtain edit tokens. The full source video is encoded into source latents, and the reference garment is encoded into garment tokens. Starting from Gaussian noise, the DiT denoises the target-video latent conditioned on source latents, garment tokens, and MLLM edit tokens. The final latent is decoded by the VAE into the edited video.

\section{Training details}
\label{app:training_details}

\subsection{Training data construction}
\label{app:training_data_construction}

Our training corpus consists of three parts: 100K image triplets, 10K ViViD video triplets, and the public TripVVT-10K training set~\citep{tripvvt}. TripVVT-10K is used as released. The other two parts are internally constructed supplementary triplets following the same source--reference--target format, where each sample contains an original person image or video, a reference garment, and a synthesized try-on target.

\paragraph{Instruction annotation.}
Each training triplet is paired with a natural-language try-on instruction. For the generated image and ViViD triplets, we use Qwen3-VL-235B-A22B-Instruct to describe the intended garment transfer for each source--garment pair. To avoid a narrow instruction distribution, we create two instruction styles for every pair: a concise command and a more descriptive request that includes additional visual context. During training, one of the two instructions is sampled with equal probability. For TripVVT-10K, where a video may contain multiple people, we first use the provided human mask to indicate the target person and then feed the marked target together with the garment reference to Qwen3-VL-235B-A22B-Instruct, so that the resulting instruction refers to the intended person rather than to the scene in general.

\paragraph{Image triplets.}
For image training data, we build candidate person--garment pairs from public image try-on datasets. Person images are taken from VITON-HD~\citep{vitonhd}, and reference garments are sampled from DressCode~\citep{dresscode}. To increase garment diversity, each person's image is paired with multiple randomly selected garments, with fewer random pairings for dress samples to avoid category imbalance. We then run CatVTON~\citep{catvton} on each candidate pair to synthesize the target try-on image, resulting in candidate triplets of the form $\langle$person image, garment image, synthesized try-on image$\rangle$.

The candidate image triplets are filtered in two stages. First, we remove samples that introduce large pose changes after synthesis. We detect 17 human keypoints in both the original and synthesized images using Sapiens~\citep{khirodkar2024sapiens}, keep keypoints with confidence above 0.3 in both images, and compute the mean normalized keypoint displacement. This displacement is converted into a pose similarity score, and only candidates with a similarity above 0.95 are kept. Second, we use a VIEScore-style evaluator~\citep{ku2023viescore} adapted to try-on data filtering. The evaluator scores perceptual quality and semantic consistency, covering clothing naturalness, artifact severity, identity preservation, and garment-transfer correctness. We combine the two scores with a harmonic mean and keep samples whose overall score is above 0.9. From the filtered pool, we retain 100K image triplets for training.

\paragraph{ViViD video triplets.}
For supplementary video data, we construct triplets from ViViD~\citep{vivid}. Because raw garment masks in video try-on data can be temporally unstable, we first filter source clips before generating synthetic targets. For each mask sequence, we compute frame-wise mask area, centroid, bounding box, normalized area change, inter-frame IoU, and centroid displacement. Frames with abrupt area changes, low temporal overlap, or large spatial jumps are marked as unstable. We also use a YOLO-based human pose detector~\citep{Jocher_Ultralytics_YOLO_2023} to check whether the mask lies in a plausible body region for the corresponding garment type, e.g., upper-body, lower-body, or dress.

Stable intervals are obtained by aggregating the temporal and anatomical checks and extracting continuous segments without detected anomalies. We keep segments that satisfy a minimum temporal length requirement and split overly long intervals when needed. Each retained person sequence is paired with a target garment, and MagicTryOn~\citep{magictryon} is used to synthesize the corresponding try-on video. We do not apply an additional generated-video filtering stage beyond the source-mask stability filtering. The final supplementary video set contains 10K ViViD triplets.

\subsection{Stage-wise training settings}
\label{app:stage_training_settings}

We summarize the reproducibility-critical protocol in prose and reserve tables for quantities that are easier to compare visually. The reported results use two public video try-on benchmarks. ViViD-S contains 180 videos released by CatV2TON, and TripVVT-Bench contains 100 videos released by TripVVT. For both benchmarks, the evaluation instructions are generated by an MLLM. Auxiliary annotations required by condition-based baselines are taken from the benchmark release when available; otherwise, we follow the corresponding method's official preprocessing pipeline.

\begin{table}[!t]
\centering
\caption{Evaluation benchmarks and inference settings.}
\label{tab:evaluation_settings}
\small
\setlength{\tabcolsep}{5pt}
\renewcommand{\arraystretch}{1.08}
\begin{tabular*}{\textwidth}{@{\extracolsep{\fill}}l c c c l@{}}
\toprule
Benchmark & Videos & Frames & Resolution & Instruction source \\
\midrule
ViViD-S & 180 & 61 & $832{\times}624$ & MLLM-generated \\
TripVVT-Bench & 100 & 49 & $832{\times}480$ & MLLM-generated \\
\bottomrule
\end{tabular*}
\end{table}

All open-source baselines are evaluated with their official checkpoints. For baselines that require masks, poses, or related structural inputs, we use their official condition-generation pipelines. Kling 1.6 is evaluated through the official web interface with manually selected edit points, matching its intended point-based interaction mode. The generated results, per-sample instructions, evaluation sample lists, metric scripts, and baseline configuration files are the key artifacts needed to reproduce the reported numbers; any public release will be reviewed against the applicable code, data, model, and asset licenses.

We train the model in four stages. The first three stages are supervised fine-tuning stages with batch size $1$, and the last stage is DiffusionNFT-based RL post-training with batch size $30$. All stages use AdamW with learning rate $1{\times}10^{-5}$. EMA is used only in Stage 3 and RL, with decay values $0.999$ and $0.5$, respectively. For the Qwen3-VL-8B conditioning MLLM in InstructVVT, the LoRA rank is $16$, and the numbers of image and video query tokens are $256$ and $512$.

\begin{table}[!t]
\centering
\caption{Training and compute settings.}
\label{tab:experimental_setting_details}
\small
\setlength{\tabcolsep}{4pt}
\renewcommand{\arraystretch}{1.06}
\resizebox{\textwidth}{!}{%
\begin{tabular}{@{}l l l c c c c c@{}}
\toprule
Stage & Objective & Data & Trainable & Steps & Batch & EMA & Hardware \\
\midrule
1 & Interface alignment & Mixed & Q, LoRA, Conn & 20K & 1 & -- & 8 H100 \\
2 & Full-condition SFT & Mixed & MLLM, DiT & 25K & 1 & -- & 8 H100 \\
3 & High-resolution SFT & TripVVT-10K & MLLM, DiT & 15K & 1 & 0.999 & 8 H100 \\
RL & DiffusionNFT & Rollouts & Gen & 180 & 30 & 0.5 & 40 H100 \\
\bottomrule
\end{tabular}%
}
\end{table}

For RL post-training, the $40$ H100 GPUs are organized as five $8$-GPU machines. On each machine, two GPUs serve the reward model and six GPUs train the policy model. Instruction annotation and reward scoring are also run on H100 GPUs. We omit absolute paths, cache locations, service ports, timeout values, and dataloader internals because they are engineering details rather than reproducibility-defining experimental settings.

\subsection{Reward model details}
\label{app:reward_details}

The reward model is a frozen Qwen3-VL-32B that is separate from the Qwen3-VL-8B conditioning MLLM. It is used only during DiffusionNFT post-training and for reporting MLLM-Reward. For each generated sample, the reward model receives the reference garment image, the original source video, the generated try-on video, and the instruction. The reward MLLM processes source and generated videos at 2 FPS. It is prompted with a fixed try-on rubric, shown in Fig.~\ref{fig:reward-prompt}, and asked to directly output a single score from 0 to 5.

For the reward-human agreement study in Sec.~\ref{sec:human_reward_eval}, each test example has four generated videos and therefore six pairwise comparisons. Pairwise agreement is the fraction of video pairs for which the reward-induced preference agrees with the human ranking. Spearman's $\rho$ measures rank correlation between the reward ranking and the human ranking, while Kendall's $\tau$ measures agreement in terms of concordant and discordant ordered pairs. These metrics are used only to validate the reward against human preference; the post-training objective uses the scalar score-token reward described below.

\begin{figure}[!t]
\centering
\includegraphics[width=0.78\linewidth]{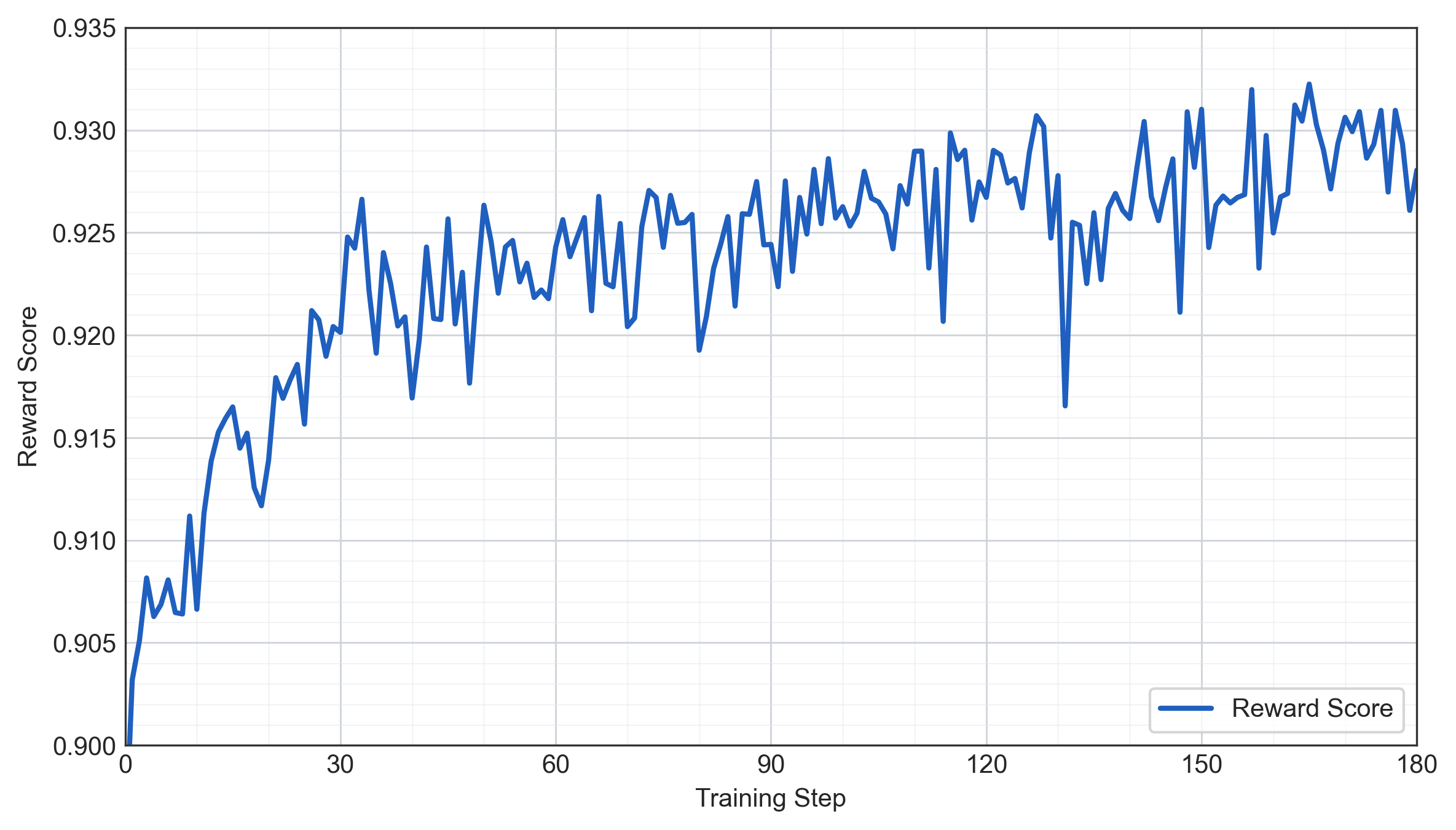}
\caption{Reward trajectory during DiffusionNFT RL post-training. The curve tracks the MLLM reward used to optimize the video generator in the RL stage.}
\label{fig:rl_reward_curve}
\end{figure}

We use a non-CoT score-token reward. Instead of parsing the generated text, we compute the expected score from the first-token logits of the six score tokens $\{0,1,2,3,4,5\}$ and normalize it to $[0,1]$, as described in Eq.~\ref{eq:score_logit_reward}. This provides a smooth scalar reward and avoids brittle text parsing.

\begin{figure}[t]
  \centering
  \fbox{\parbox{0.95\linewidth}{\small
  \textbf{System prompt for video try-on reward scoring.}\\[3pt]
  Human: You are evaluating a video virtual try-on result.\\[3pt]
  You will receive: \\
  1.\ A reference garment image.\\
  2.\ The original/source video.\\
  3.\ The generated try-on video.\\[3pt]
  Instruction: \{\texttt{prompt}\}\\
  Requirements: \{\texttt{requirement}\}\\[3pt]
  Rate the generated try-on video from 0 to 5 based on the following criteria.\\[3pt]
  (1) Target Clothing Replacement\\
  -- The clothing on the specified target body region should be replaced with the reference garment.\\
  -- If the model edits the wrong person, wrong region, or does not apply the garment, the score should be low.\\[3pt]
  (2) Reference-Garment Fidelity\\
  -- The replaced clothing should match the reference garment in category, color, texture, pattern, shape, and visible design details.\\
  -- Major mismatch with the reference garment should be penalized.\\[3pt]
  (3) Source Preservation\\
  -- Human regions outside the edited clothing area should be preserved, including identity, pose, body shape, face, hands, skin, and background.\\
  -- Unnecessary changes to non-edited regions should be penalized.\\[3pt]
  (4) Visual Quality and Temporal Stability\\
  -- The generated video should be visually natural, temporally stable, and free of severe artifacts.\\
  -- Flickering clothing, inconsistent details, or severe distortions should be penalized.\\[3pt]
  Use the following scale:\\[2pt]
  0: The try-on fails, edits the wrong target region, ignores the reference garment, or has severe artifacts.\\
  5: The specified clothing region matches the reference garment accurately, all non-edited human regions and the background are preserved, and the video is visually high quality.\\[3pt]
  Response Format: directly output one score number from 0 to 5.\\[2pt]
  Do not output anything else.
  }}
  \caption{System prompt used for MLLM reward scoring.}
  \label{fig:reward-prompt}
\end{figure}
\subsection{Other implementation details}

We use a fixed negative prompt for Wan-style video sampling and keep it identical across supervised and RL stages. Unless otherwise specified, all RL experiments use random seed 42.

\section{Human evaluation details}
\label{app:human_eval_details}

The human preference study uses $50$ TripVVT-Bench examples and four methods: Ours, MagicTryOn, UniVideo, and Kling 1.6. Each evaluation task shows the source video, the reference garment, the instruction, and four anonymized generated videos in randomized order. Participants are asked to rank the four results according to overall video try-on quality, considering whether the instructed target clothing is replaced by the reference garment, whether the source identity, body shape, motion, and background are preserved, whether the garment appearance is temporally stable, and whether the video is visually natural. We collect only preference rankings and do not collect personally identifying information from participants.

The instruction shown to participants is:
\begin{quote}
You will see a source video, a reference garment, an editing instruction, and four anonymized generated videos. Please rank the four generated videos from best to worst according to overall video virtual try-on quality. A better result should apply the reference garment to the instructed target person, preserve non-edited identity, pose, motion, and background content, maintain temporal consistency, and avoid visual artifacts.
\end{quote}

The reward-human agreement study follows the same ranking interface and criteria, but is used to compare human rankings with reward-induced rankings rather than to compute the four-method first-place vote ratios. Participants are informed that the study evaluates generated video try-on results for research purposes and can stop the evaluation at any time. Compensation is handled under the authors' internal annotation policy and is independent of the submitted rankings.

\section{Additional analysis of metric discrepancies}
\label{app:metric_discrepancy}

\paragraph{ViViD-S metric behavior.}
The ViViD-S results reveal an important limitation of generic video quality metrics for try-on evaluation. MagicTryOn performs strongly on VFID, CLIP-F, and background metrics, outperforming ViViD and CatV2TON on several low-level or distribution-level measures. However, its MLLM-Reward is substantially lower than those of ViViD, CatV2TON, and TripVVT. Qualitative inspection suggests that ViViD-S is mostly composed of relatively simple indoor videos, where many methods can preserve the person and background reasonably well; under this setting, a small number of severe target-clothing replacement failures can dominate try-on-specific scoring, because the reward penalizes wrong or incomplete garment transfer more strongly than generic video metrics.

\paragraph{UniVideo on ViViD-S.}
UniVideo shows a different behavior. As a general instruction-guided video editor, it achieves very high CLIP-F and a high MLLM-Reward, suggesting that it can often understand and execute the requested edit. Nevertheless, its VFID$_I$, VFID$_R$, BG-L1, and BG-DINO are much worse than specialized try-on methods. This indicates that general video editing ability alone is insufficient for video virtual try-on: the model may follow the instruction, but still fail to preserve the source distribution, background, and local editing constraints required by high-quality try-on.

\paragraph{Commercial editors on TripVVT-Bench.}
The comparison with commercial editors further highlights the difficulty of video virtual try-on. Kling 1.6 obtains a relatively strong MLLM-Reward among non-specialized models, which is consistent with its point/region-based editing interface: the additional region-level hint provides useful localization information, similar in spirit to the auxiliary spatial priors used by previous try-on systems. However, it still lags behind InstructVVT in garment fidelity, reconstruction quality, and human preference. Kling O3, despite being a more recent general-purpose multimodal video generation system, performs substantially worse on this benchmark. Qualitative inspection suggests that Kling O3 tends to regenerate the video according to the input reference and instruction rather than strictly preserving the source layout and performing a localized clothing edit. These results show that even strong general video generation models do not automatically solve video virtual try-on, and that recovering try-on-specific controllability without inference-time structural priors remains necessary.

\section{Additional ablation analysis}
\label{app:ablation_analysis}

The ablation variants are designed to isolate the major condition pathways in InstructVVT. Replacing MLLM conditioning with the original T5 text encoder tests whether MLLM-based conditioning provides useful video- and garment-aware editing context beyond text-only conditioning. Removing the reference garment from the MLLM input while keeping the generator-side garment tokens isolates the contribution of giving the MLLM access to the reference garment for garment-to-person alignment. Removing the source-video condition tests whether the input video latent is necessary for preserving motion, layout, and non-edited regions. Removing generator-side reference-garment tokens by zeroing the reference-garment latent tests whether the garment token branch contributes fine-grained reference appearance beyond the reference information already available to the MLLM. Finally, removing DiffusionNFT-based post-training tests whether MLLM reward post-training improves try-on-specific qualities beyond supervised fine-tuning.

The results in Table~\ref{tab:ablation} support this decomposition. Replacing the MLLM with T5 leads to worse performance, especially on metrics related to target correctness, source preservation, instruction following, and garment-reference alignment. Removing the garment input from the MLLM weakens garment-reference alignment while retaining the generator-side garment tokens, indicating that access to the reference garment helps the MLLM provide high-level guidance on how the garment should be applied to the target person. Removing the source-video condition harms preservation and temporal stability, confirming that source latents are important anchors for the original motion, layout, and background. When generator-side reference-garment tokens are removed, the MLLM can still provide garment-aware intent, but reference-specific fidelity and fine details degrade. Removing DiffusionNFT-based post-training also degrades performance, indicating that the proposed MLLM reward provides useful preference feedback for improving garment fidelity, preservation, and temporal consistency. The full model improves the reward score and most independent metrics, suggesting that the gain reflects broader try-on quality improvement rather than only reward-score optimization.

\section{Broader impacts and safeguards}
\label{app:broader_impacts}

Instruction-driven video virtual try-on can support online retail, digital content creation, and accessibility by reducing the need for repeated physical try-on or manual video editing. At the same time, the same capability may be misused to alter a person's appearance without consent, create misleading fashion or identity-related media, or generate edited videos that viewers may mistake for authentic recordings. These risks are especially relevant for in-the-wild videos that contain identifiable people.

Responsible deployment should therefore require consent from depicted individuals, restrict use on sensitive or non-consensual imagery, and clearly disclose generated or edited content. If models, demos, or generated data are released, we plan to include usage restrictions, safety filtering for inappropriate inputs, and documentation of intended and prohibited uses. We also encourage downstream systems to combine try-on generation with provenance or watermarking mechanisms where feasible.

\section{Asset licenses and terms}
\label{app:asset_licenses}

We use external assets for data construction, benchmark evaluation, baseline comparison, filtering, annotation, reward modeling, and generator initialization. Table~\ref{tab:asset_licenses} lists only the asset category, role, and public license or terms status. When a release does not specify a standard open-source license, we report the official terms status rather than inferring one.

\begin{table}[!t]
\centering
\caption{External assets used in this work and their public license or terms status.}
\label{tab:asset_licenses}
\footnotesize
\setlength{\tabcolsep}{8pt}
\renewcommand{\arraystretch}{1.05}
\resizebox{\textwidth}{!}{%
\begin{tabular}{@{}llll@{}}
\toprule
Asset & Category & Role & License / terms \\
\midrule
VITON-HD & Dataset & Image source persons & CC BY-NC 4.0 \\
DressCode & Dataset & Reference garments & Academic non-commercial terms \\
ViViD & Dataset / benchmark & Video triplets; ViViD-S & Apache-2.0 \\
TripVVT-10K / TripVVT-Bench & Dataset / benchmark & Training/evaluation & Official release terms \\
CatVTON & Method / baseline & Image target synthesis & CC BY-NC-SA 4.0 \\
MagicTryOn & Method / baseline & Video target synthesis & CC BY-NC-SA 4.0 \\
UniVideo & Method / baseline & General video editing baseline & MIT \\
Sapiens & Tool & Keypoint filtering & CC BY-NC 4.0; Apache-2.0 portions \\
Ultralytics YOLO & Tool & Pose/body-region checks & AGPL-3.0 or Enterprise \\
Qwen3-VL-8B / 32B / 235B-A22B-Instruct & Model & Condition; reward; annotation & Apache-2.0 \\
Wan2.1-style backbone & Model & Generator initialization & Apache-2.0 \\
Kling 1.6 / Kling O3 & Service & Commercial baseline & Provider service terms \\
\bottomrule
\end{tabular}%
}
\end{table}

Assets governed by non-commercial licenses, academic-use agreements, or service terms are used only within the corresponding research, evaluation, or service-access context. Any public release of derived data, generated results, checkpoints, or code will be reviewed against the applicable asset terms.

\begin{figure}[!p]
\centering
\makebox[\textwidth][c]{\includegraphics[width=1.04\textwidth]{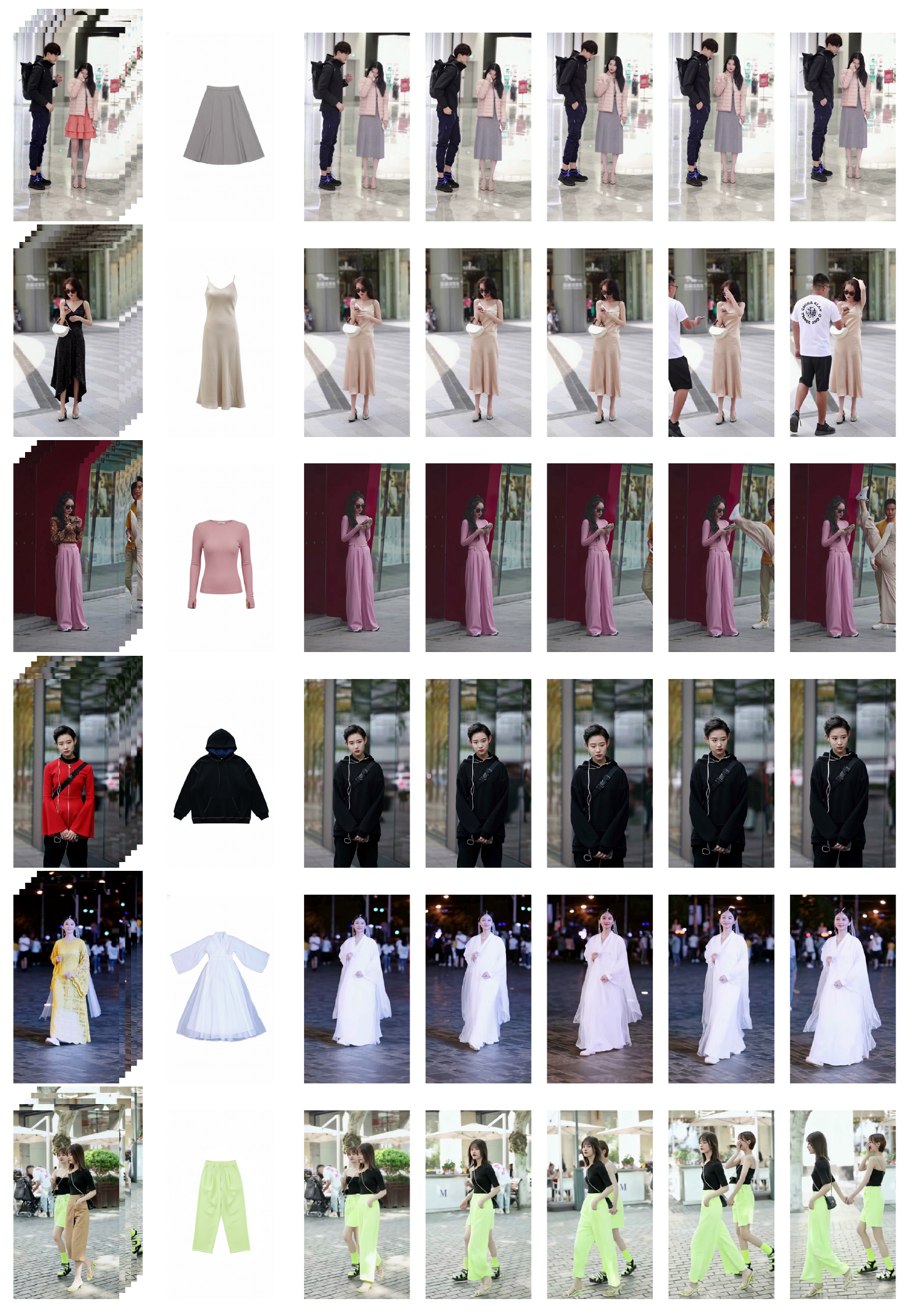}}
\caption{Additional qualitative comparisons on TripVVT-Bench. The examples cover diverse in-the-wild scenes, reference garments, and target-person configurations, complementing the main-paper qualitative results.}
\label{fig:additional_qualitative_tripvvt}
\end{figure}

Additional qualitative comparisons on TripVVT-Bench are shown in Fig.~\ref{fig:additional_qualitative_tripvvt}. These examples supplement the main-paper visual comparison by covering more in-the-wild scenes, reference garments, and target-person configurations.

\clearpage
\section{Additional qualitative results}
\label{app:additional_qualitative_results}

Additional qualitative comparisons are provided for both ViViD-S and TripVVT-Bench. ViViD-S visualizations complement Table~\ref{tab:vivid_s} by showing frame-level garment transfer and preservation behavior on the controlled benchmark, while TripVVT-Bench visualizations cover more in-the-wild scenes and target-person configurations. We also include a qualitative ablation comparison to illustrate how the major conditioning pathways affect the generated videos.

\begin{figure}[!htbp]
\centering
\includegraphics[width=0.68\textwidth]{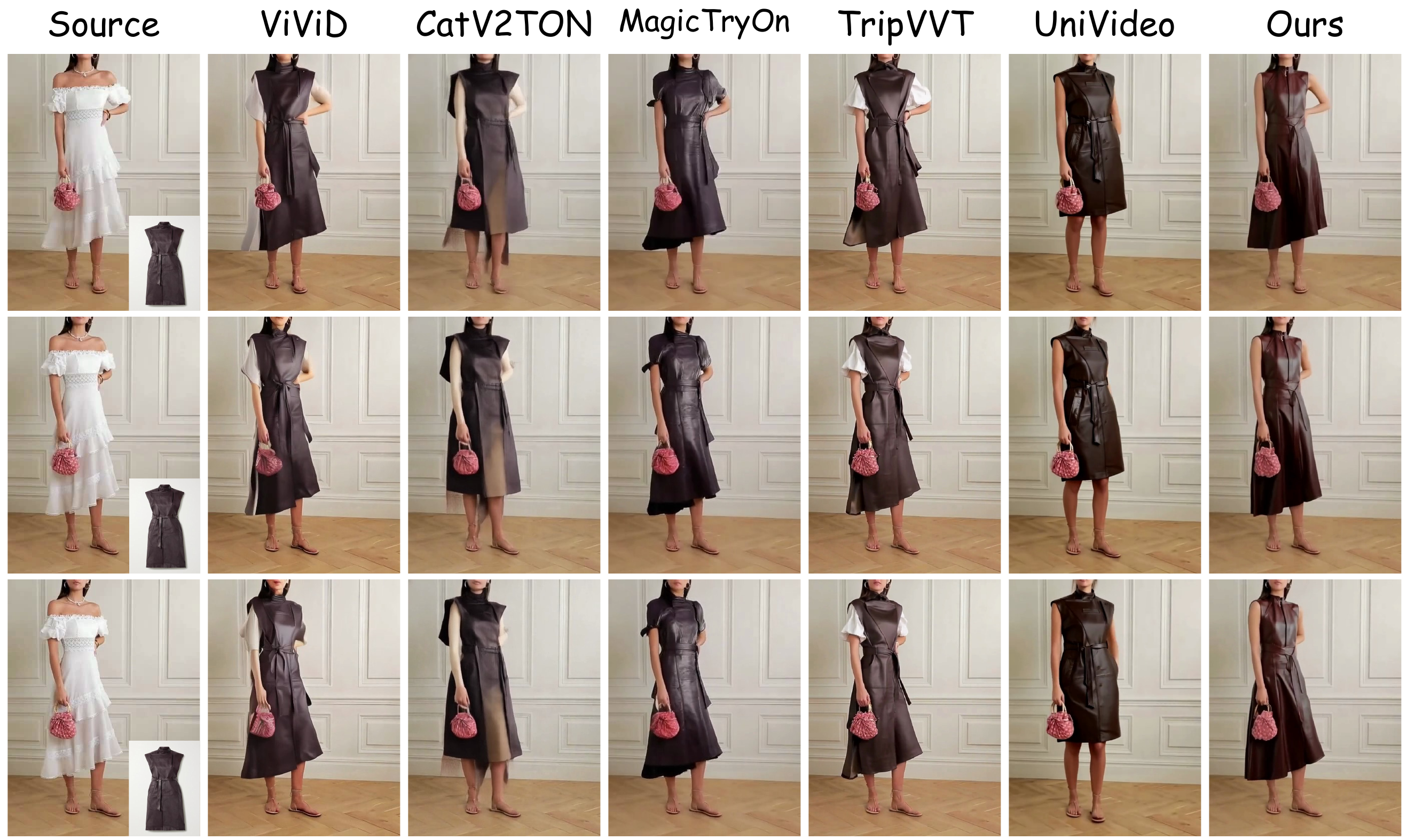}
\caption{Qualitative comparisons on ViViD-S. The examples show that InstructVVT transfers the reference garment while preserving the source video appearance and temporal structure, complementing the quantitative ViViD-S results in Table~\ref{tab:vivid_s}.}
\label{fig:additional_qualitative_vivid_s}
\end{figure}

\begin{figure}[!htbp]
\centering
\includegraphics[width=0.82\textwidth]{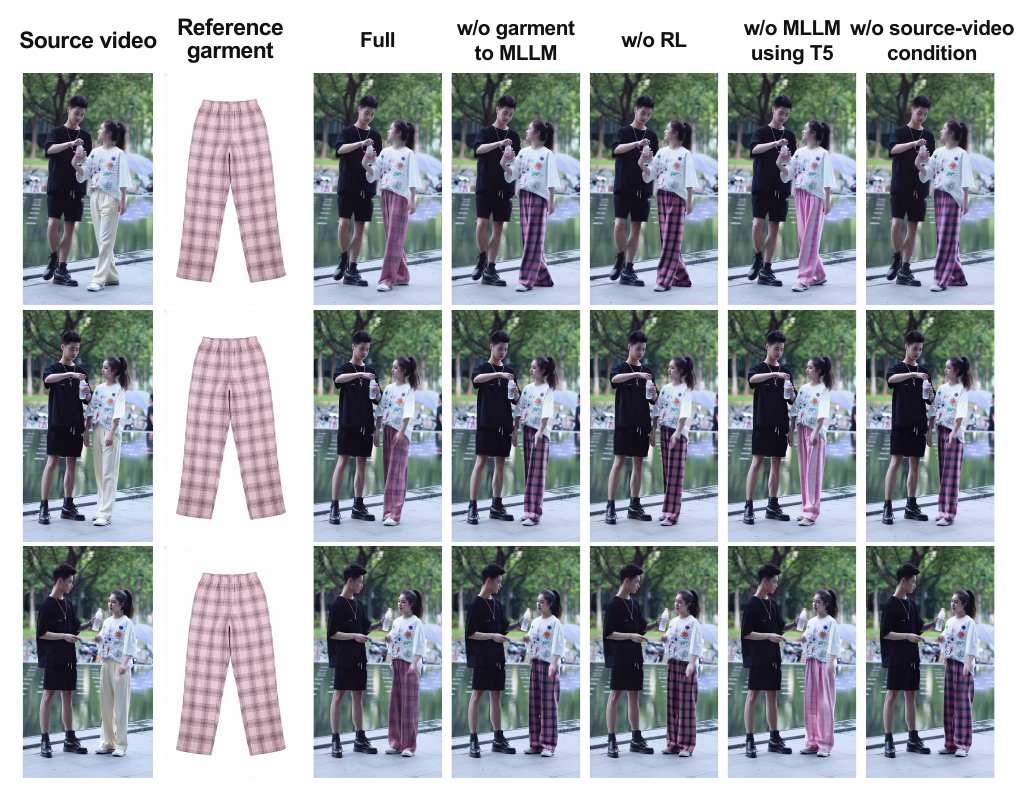}
\caption{Qualitative ablation comparison on TripVVT-Bench. Replacing the MLLM with T5 weakens video- and garment-aware edit conditioning; removing the reference garment from the MLLM input reduces high-level garment-to-person alignment; removing source-video conditioning weakens preservation; and removing DiffusionNFT post-training reduces the overall try-on quality. These visual trends are consistent with the quantitative ablations in Table~\ref{tab:ablation}.}
\label{fig:additional_ablation_visual}
\end{figure}

\begin{figure}[!p]
\centering
\makebox[\textwidth][c]{\includegraphics[width=1.08\textwidth]{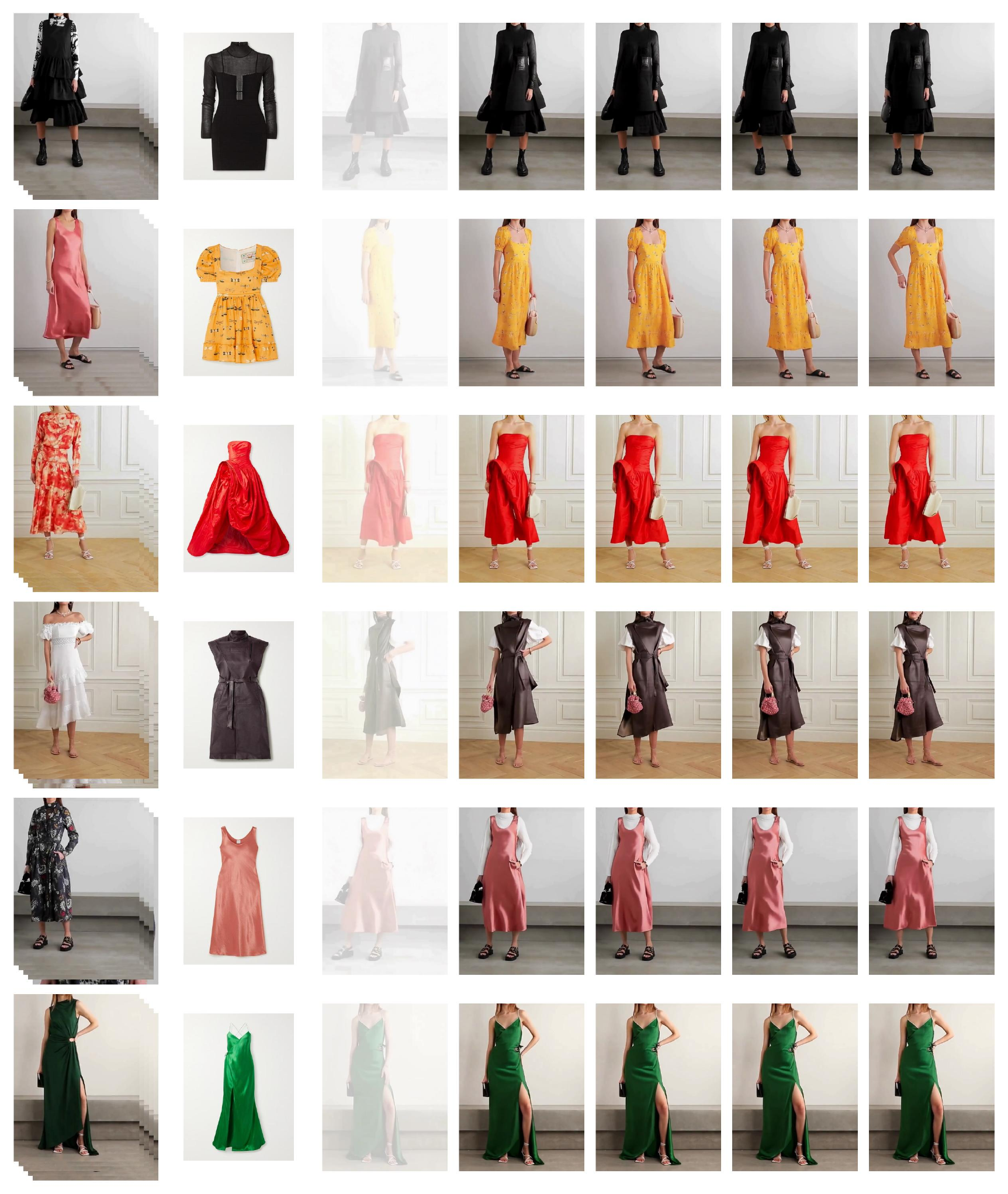}}
\caption{Additional ViViD-S visualizations. These samples provide a broader view of the method's behavior across different source videos and reference garments, showing stable garment appearance and source preservation across sampled frames.}
\label{fig:additional_qualitative_vivid_s_more}
\end{figure}

\end{document}